\documentclass{article} 
\usepackage{nips13submit_e,times}
\usepackage{hyperref}
\usepackage{url}
\usepackage{amsmath}
\usepackage{graphicx}

\title{A Non-parametric Conditional Factor Regression Model for High-Dimensional Input and Response}

\author{
Ava Bargi\\
University of Technology, Sydney \\
\texttt{ava.bargi@uts.edu.au} \\
\And
Richard Yi Da Xu \\
University of Technology, Sydney \\
\texttt{YiDa.Xu@uts.edu.au} \\
\And
Massimo Piccardi \\
University of Technology, Sydney \\
\texttt{Massimo.Piccardi@uts.edu.au} \\
}

\nipsfinalcopy 

\begin{document}

\maketitle

\begin{abstract}
In this paper, we propose a  non-parametric conditional factor regression (NCFR) model for domains with high-dimensional input and response. NCFR enhances linear regression in two ways: a) introducing low-dimensional latent factors leading to dimensionality reduction and b) integrating an Indian Buffet Process as a prior for the latent factors to derive unlimited sparse dimensions. Experimental results comparing NCRF to several alternatives give evidence to remarkable prediction performance.
\end{abstract}

\section{Introduction}
With the exponential growth in data generation, multi-variate problems with high-dimensional input and output are becoming more common. Volatility matrix estimation and price forecasting in finance, as well as action prediction, view-to-view recognition \cite{Quang_gateCVPR10} and pose estimation \cite{bo2009supervised} in computer vision are examples of such problems with high-dimensional features. Linear regression has been a long-standing, simple but effective prediction tool for many domains. Yet, most of the existing solutions cater for one-dimensional response, leaving a visible gap for multi-variate high-dimensional cases. In this paper, we propose an elaborate regression model to cater for such prediction contexts, exploring a few variations for improvement.

Let us begin with the classic regression model $Y = RX + E$ in high dimensions, where $Y$ is the $q$-dimensional response over n observations ($q \times N$) and $X$ is the $p \times N$ input, regressed by $R$ and added diagonal Gaussian noise. For large $p$ and $q$, $R$ would be a huge matrix imposing matrix multiplications of order $O(qpN$), that are both computationally costly and could be numerically inaccurate.

A step forward, we can improve the model through introducing a latent factor, $Z$, not only bridging $X$ and $Y$, but jointly reducing their ranks to $K << p,q$. Such latent factors also improve the noise model by decoupling it for input and response into separate $\epsilon_z$ and $\epsilon_y$ \cite{west2003bayesian}. Let us call this parametric model  {\it conditional factor regression} (CFR). Variants of such models are studied in \cite{west2003bayesian} and \cite{bo2009supervised}, but are subject to a prior decision making on the optimal latent dimensionality ($K$) through trial and error or domain knowledge. 

To resolve this, we propose a {\it `non-parametric' conditional factor regression} (NCFR) model: a novel Bayesian non-parametric treatment to multi-variate high-dimensional regression that facilitates finding an optimal $K$-dimensional latent layer by exploiting an Indian Buffet Process (IBP) prior \cite{IBPReport2005}. In section 2, we explore the background research on similar models, followed by an articulate description of NCFR model parameters and inference in sections 3 and 4. Through experiments in section 5, we evaluate and compare the above-mentioned models, following with the Conclusion. 

\section{Background}
\label{sec:background}

The notion of latent factor regression has been introduced with varying terminology and design. Bayesian factor regression model \cite{west2003bayesian} represents a regression model, particularly proposed for large input size and small observation number. As the name implies, it uses a latent factor between input and response. As opposed to the proposed NCFR, the response variable is one-dimensional and both input and response are dependent on latent factors. The spectral latent variable model proposed in \cite{bo2009supervised}, however, shares the same design as ours, creating a transitive dependency of response over the latent factor, and in turn over the input (\ref{fig:GraphicalModel}). The dependencies between high-dimensional inputs and responses are channelled via a low-dimensional latent manifold using mixtures of RVMs.

Both the above models require a parametric choice of $K$ (dimensionality of latent layer). Caron and Doucet have remedied this through proposing a class of priors based on scale mixture of Gaussians over the regressor \cite{caron2008sparse}. Their sparse Bayesian non-parametric model is essentially an infinite sparse regressor, correlating an infinite dimensional input to a single dimensional response through Levy processes. However, this solution is not computationally efficient for high-dimensional regression. 
Alternatively, one can utilise an IBP prior to select the relevant factors for each observation through a sparse prior. Infinite sparse factor models exploit this property of IBP to enhance factor analysis, through sparsifying either the factors \cite{knowles2007infinite} or the load factor matrix \cite{knowles2011nonparametric}, \cite{rai2009multi}. Our model follows the former by imposing sparsity over the latent factors ($Z$), yet cascading conditional dependence of $Y$ on $Z$ and transitively over $X$.

\subsection{Indian Buffet Process: the binary mask}

The art of an Indian buffet process is that of constructing an infinite sparse binary matrix, $S$. Griffith and Ghahramani \cite{IBPReport2005} initially introduce a finite case for $S$, consisting of $K$ features and further extend it to the asymptotic infinite version where $K \rightarrow \infty$. Each feature $k$ is active with a Binomial likelihood $\pi_k$, integrated out for simplicity. The resulting density for $S$ is a normalised Poisson distribution over $\alpha$
\begin{equation}
\begin{split}
P(S) = { \alpha^{K}  \over \prod_{h>0} K_h !}  exp(- \alpha H_N)  \prod_{k=1}^{K} {{(N - m_k)! (m_k - 1)! \over N!}}\\
\end{split}
\end{equation}
where $\alpha$ is the strength parameter and $m_k = \sum_{n=1}^N{s_k}$ is the number of data points for which feature $k$ is active. $H_N = \sum_{j=1}^N{1 \over j}$ is the $N$-th harmonic number and $K_h$ is the number of rows whose binary values are equal to the decimal number $h$.

Using the above distribution, the Indian buffet analogy is phrased as a buffet with an infinite number of dishes allowing customers to try unlimited options. The probability of choosing each dish is either driven by previous customers' choices, or drawn from a $Poisson$ prior for untried dishes. In other words, each customer $i$ will select an already tried dish with probability $m_k \over i$, and/or decide to try a new dish with probability $Poisson({\alpha \over i})$. Thanks to exchangeability, each customer $i$ could be treated as the last; thus, informing each decision with {\it all} other choices ($s_{-ki}$). 
\begin{equation}
\label{eq:prior}
\begin{split}
&P(s_{ki} = 1 | s_{-ki}) = {m_{k,-i} \over N}, \hspace{1cm}  m_{k,-i} = \sum_{n \neq i} s_{kn}\\
\end{split}
\end{equation}
\section{The model}
\begin{figure}
\begin{center}
\includegraphics[scale = 0.7,  clip]{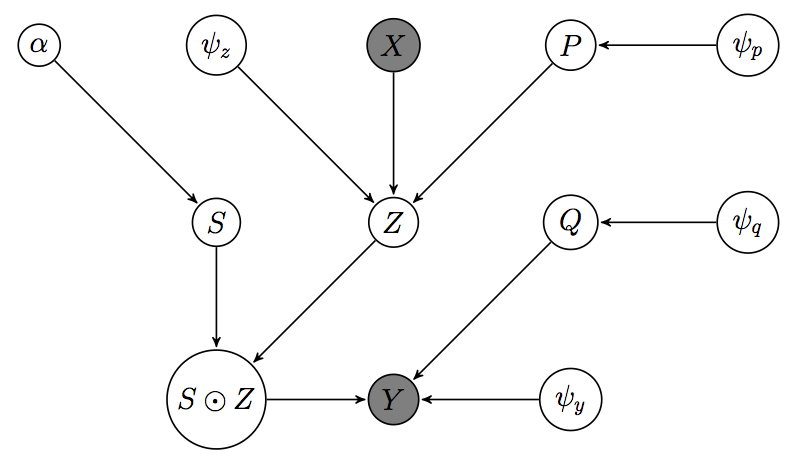}
\end{center}
   \caption{NCFR Graphical model.}
\label{fig:GraphicalModel}
\end{figure}

Non-parametric Conditional Factor Regression (NCFR) consists of two linear Gaussian transforms, linked through a sparse latent layer:
\begin{equation}
\label{eq:model}
\begin{split}
p(Y|Z)= N(Y|QZ, \Psi_y), \\
p(Z|X)= N(Z|PX,\Psi_z) \\ 
\end{split}
\end{equation}
assuming $X$, $Y$ and $Z$ consist of $N$ i.i.d. observations. To maintain reasonable generality, we have considered diagonal covariances also reserving the option for isotropic variances in simpler models. As mentioned earlier, an important challenge remains selecting the optimal dimensionality of $Z$, which yields the best regression and dimension reduction performance. In conventional cases, this is a rigid decision. However, by adding an IBP prior to the model, we have derived an infinite sparse $Z$ to best fit the data. Thus, the model is defined as follows:
\begin{equation}
\label{eq:model2}
\begin{split}
Y = Q(S \odot Z) + E_y, \\
(S \odot Z) = P X + E_z \\
\end{split}
\end{equation}
where $S$ is the IBP binary mask applied to $Z$, determining whether or not each dimension is active. The weight of each active feature in $S$ is specified via its respective entry in $Z$. 
Ultimately, each of the p-dimensional input vectors, $x_n$ in $X_{p \times N}$, are regressed into a respective $y_n$ response vector of q dimensions ($Y_{q \times N}$), via sparse $K$-dimensional $z_n$s that are each masked through a relevant binary $s_n$ through element-wise Hadamard product $(S \odot Z)_{K \times N}$. The result is summed with Gaussian noise terms, sampled from diagonal covariances $\Psi_y$ and $\Psi_z$. The noise terms collectively form $E_{y (q \times N)}$ and $E_{z (p \times N)}$.  $Q$ and $P$ are load factor matrices, comprised of independent Gaussian vectors with diagonal covariances $\Psi_q$ and $\Psi_p$. The binary mask matrix $S$ is sampled from an IBP prior with Gamma-distributed $\alpha$. Figure \ref{fig:GraphicalModel} illustrates the proposed graphical model.

\vspace{0.2cm}

\begin{equation}
\label{eq:assumptions}
\begin{split}
&E_{y} \sim N(0, \Psi_y), \hspace{0.3cm}    \Psi_{y}(i)  \sim IG(a,b),  \hspace{1.5cm}
E_{z} \sim N(0, \Psi_z), \hspace{.3cm}    \Psi_{z}(k)  \sim IG(a,b), \\
&P \sim N(0, \Psi_{p}),  \hspace{0.5cm}  \Psi_{p}(k)  \sim IG(c,d), \hspace{1.5cm}
Q \sim N(0, \Psi_{q}),  \hspace{0.3cm}  \Psi_{q}(k)  \sim IG(c,d), \\
&S \sim IBP(\alpha), \hspace{0.7cm}  \alpha \sim G(g,h) \\
\end{split}
\end{equation}

\section{Inference}
Given the input and response observations $X$ and $Y$, we tend to infer model's random parameters jointly in a posterior probability, using Gibbs sampling. Additional Metropolis-Hastings(MH) steps are utilised particularly for deriving the newly activated features in $S$ \cite{knowles2007infinite}. This section explains the inference details for all random variables introduced in \eqref{eq:assumptions}.

\subsection{Binary mask}
Complying with the definition of IBP and similar infinite processes, $S_{K \times N}$ is a matrix with an infinite number of sparse rows. For simplicity and memory efficiency, we tend to only consider the active features ( $\exists \hspace{1mm} n \hspace{1mm}| \hspace{1mm} s_{kn} =1$) and sample their related weights. Thus, $S$ could extend or shrink in each iteration, due to customers' choices. In this section, we begin with sampling the existing active features and proceed with adding new ones. 

To sample the binary elements of $S$, we form the ratio of posteriors for active vs inactive $s_{kn}$ elements. Thanks to this ratio, normalization factors are canceled out and a uniform random draw can determine whether or not $s_{kn}$ is active. 

\begin{equation}
\begin{split}
&r_l = {N_l \over D_l} = {P(y_n|s_{kn} = 1, s_{-kn}, z_{-kn}, Q, \Psi_y) \over P(y_n|s_{kn} = 0, s_{-kn}, z_{-kn}, Q, \Psi_y)}, \\
&r_p = {P(s_{kn} = 1 | s_{-kn}) \over P(s_{kn} = 0 | s_{-kn})}, \hspace{1.7cm} r = r_l . r_p \\
\end{split}
\end{equation}

Before attempting to derive this ratio, the likelihood terms are marginalised with respect to $z_{kn}$, the latent factor weight of $k$-th feature in the $n$-th observation. However, these weights are sampled for all active $s_{kn}$. Following Gaussian linear transformation properties, the resulting marginal density is Gaussian with known parameters, provided that the likelihood mean is represented as a linear transformation over $z_{kn}$ ($A z_{kn} + b$). Using ordinary matrix algebra, we have decomposed the mean $Qz_n$ into two terms. The former includes the parameter of interest $z_{kn}$, whereas the latter (residue term) excludes $z_{kn}$ through a reduced product $[Qz_n]_{z_{kn}=0}$ with one less element in $z_n$. Yet, to maintain the dimensional compatibility, we have kept the $k$-th element and made it equal to zero. Hence, the marginal likelihood for the active case is distributed as follows:

\begin{equation}
\label{eq:Nl}
\begin{split}
&N_l = \int{N(y_n|Qz_n, \Psi_y)N(z_{kn}|p_kx_n, \Psi_z)dz_{kn}}, \hspace{3mm} Qz_n = q_k z_{kn} + [Qz_n]_{z_{kn} = 0} \\
& \hspace{1cm} = N (y_n | q_k p_k x_n + [Q z_n]_{z_{kn} = 0} , \Psi_y + q_k {\Psi_z(k)}^{-1} q_k^T) \\
\end{split}
\end{equation}

In case of inactive $s_{kn}$, the Hadamard product $s_{kn} \odot z_{kn}$ is inevitably zero for every $z_{kn}$. Therefore, the prior probability for marginalisaton is no longer informative, resulting in $D_l$ below. The likelihood ratio $r_l$ could be achieved from \eqref{eq:Nl} and \eqref{eq:Dl}. 

\begin{equation}
\label{eq:Dl}
\begin{split}
&D_l = N(y_n|[Qz_n]_{z_{kn} = 0}, \Psi_y) \\ 
\end{split}
\end{equation}

The prior ratio $r_p$ can be derived as $r_p = { m_{k,-i} \over N - 1 - m_{k,-i}}$, following \eqref{eq:prior}. The $i$-th observation is not counted for in the ratio, which justifies the $-1$ in the denominator. Having derived $r_l$ and $r_p$, the posterior ratio for existing features could be derived as a uniform draw over $\frac{r}{r+1}$ . 

Next, we should decide the number of new features ($\kappa_n$) added for the current observation. IBP implies that $\kappa_n$ is {\it a priori} distributed as $Poisson(\frac{\alpha}{n})$ \cite{IBPReport2005}. Yet, there have been different approaches for sampling $\kappa_n$ from its posterior. In \cite{IBPReview2011} $\kappa_n$ is sampled through a MAP estimate of various values for $\kappa_n$, ranging from zero to an upper bound. However, \cite{knowles2011nonparametric} proposes a Metropolis-Hastings step to some random $\kappa_n$ and evaluates the acceptance through posterior ratios. Each MH jump from $\xi \rightarrow \xi^*$ is done with a probability $J(\xi^*|\xi)$ with varying underlying assumptions. A basic approach considers the prior on $\xi^*$ as candidate function \cite{knowles2007infinite}. As a more elaborate alternative, \cite{meeds2006alternative} introduced a spike and slab prior to create bias in favour of $\kappa_n = 1$ . As a special case, we have tried a zero function to allow absolutely no new feature, confining IBP into a fixed sparse algorithm. We have also proposed an additional Simulated Annealing weight to encourage fewer new factors as the iterations proceed and ultimately regulate model's fitting over time. The discussion in section 5 presents a comparison amongst the above alternatives. 

\begin{equation}
\label{eq:MHratio}
\begin{split}
&r_{\xi \rightarrow \xi^*} = {P(\xi^* | y_n, \_)J(\xi|\xi^*) \over P(\xi | y_n, \_)J(\xi^*|\xi)} = {P(\xi^* | y_n, \_)e^{-\kappa_n \over T}P(\xi^*)P(\xi)  \over P(\xi | y_n, \_)P(\xi)P(\xi^*)}\\
&r_{\xi \rightarrow \xi^*} = e^{-\kappa_n \over T}{N_r \over D_r} = e^{-\kappa_n \over T}{P(\xi^* | y_n, \_) \over P(\xi | y_n, \_)}\\
\end{split}
\end{equation}

In the equations above, $\_$ denotes the remaining parameters and * superscripts indicate extended parameters with $\kappa_n$ new features. Simulated annealing assumption implies that the probability of choosing $\kappa_n$ new features, $J(\xi^*|\xi)$, is simplified to $e^{-\kappa_n \over T}P(\xi^*)P(\xi)$ where the exponential weight is gradually decreased through declining temperature $T$, thereby decreasing the acceptance probability for adding features. The jump is ultimately accepted with probability $min (1, r_{\xi \rightarrow \xi^*})$.

In order to add $\kappa_n$ new features, we need to add an equal number of new columns and rows to $Q$ and $P$, respectively. These new vectors are denoted with prime notations as in $\xi^* = \{\kappa_n, q^\prime,p^\prime\}$. Please note that along with activating $\kappa_n$ new features, an equivalent number of weight elements ($z^\prime$) is needed. Such weights are marginalised for simplicity. 

\begin{equation}
\label{eq:Zmarginalized}
\begin{split}
&N_r = N (y_n | q_k^\prime p_k^\prime x_n + [Q^* z_n]_{z_{kn}^\prime = 0} , \Psi_y + q_k^\prime {\Psi_z(k)}^{-1} q_k^{\prime^T}) \\
&D_r = N(y_n|[Q z_n]_{z_{kn}^\prime = 0}, \Psi_y)\\
\end{split}
\end{equation}

\subsection{Load factor matrices}
Inference in load factor matrices can be uniformly described for new and existing features, through sampling $K$ independent vectors in $Q$ and $P$. Following the graphical model (figure \ref{fig:GraphicalModel}), the load factor matrices are independent from each other through $Z$. Thus, their posterior probabilities can be independently derived. As follows, we infer column vectors $q_k$ through sampling individual elements $q_k(i), i \in [1,q]$. $QZ$ is linearly decomposed into $ q_{k}z_{k} + [QZ]_{q_{k} = 0}$ and transposed to derive the following posterior (eq. \ref{eq:Q}). Following a similar approach, we propose the posterior for $p_{k}$ (eq. \ref{eq:P}).  

\begin{equation}
\label{eq:Q}
\begin{split}
& N(q_k(i)|\mu_{q_k(i)},\Sigma_{q_k(i)}) \propto  N(y_i|QZ, \Psi_y(i)) N(q_{k}(i)| 0, \Psi_{q}(i)I_q) \\
&\Sigma_{q_k(i)} = {\Psi_q(k) \Psi_y(i) \over \Psi_y(i) + \Psi_q(k) z_k z_k^T}, \hspace{3mm} 
\mu_{q_k(i)} = {\Psi_q(k) (y_i - [QZ]_{q_{ik} =0}) z_k^T \over \Psi_y(i) + \Psi_q(k) z_k z_k^T}\\
\end{split}
\end{equation}

\begin{equation}
\label{eq:P} 
\begin{split}
&N(p_{k}| \mu_{p_{k}}, \Sigma_{p_{k}}) \propto  N(z_k|p_kX, \Psi_z(k)) N(p_{k}| 0, \Psi_{p}(k)I_p) \\
&\Sigma_{p_{k}} =  \Psi_{p}(k)I_p +  \Psi_{z}(k) (X X^T)^{-1}, \hspace{3mm} \mu_{p_{k}} =  \Sigma_{p_{k}} {z_k X^T \over \Psi_z(k)}\\
\end{split}
\end{equation}

\subsection{Latent factor weights}
Each active feature $s_{kn}$ is assigned a weight ($z_{kn}$), the element-wise product of which forms the infinite sparse latent layer between the high dimensional input ($X$) and response ($Y$). The posterior on each $z_{kn}$ can be derived through a Bayesian treatment \cite{Bishop2006}:

\begin{equation}
\label{eq:postZ}
\begin{split}
& N (z_{kn}|\mu_{z_{kn}}, \Sigma_{z_{kn}} ) \propto N(Y | q_{k}z_{kn} + [QZ]_{z_{kn} = 0}, \Psi_y)N(z_{kn}|P_{k}x_n, \Psi_z(k)) , \\
& \Sigma_{z_{kn}} = ({1 \over \Psi_z(k)} + {q_k^T \Psi_y^{-1} q_k}) ^ {-1}, \hspace{3mm} \mu_{z_{kn}} = \Sigma_{z_{kn}}[q_k^T \Psi_y^{-1} (Y - [QZ]_{z_{kn} = 0}) + {p_k x_n \over \Psi_z(k)}] \\
\end{split}
\end{equation}

\subsection{Noise, loading factors and IBP parameters}
NCFR adopts diagonal noise models, $\Psi_y$ and $\Psi_z$. The loading factors' covariances, $\Psi_q$ and $\Psi_p$, are also diagonal. Following \cite{knowles2011nonparametric}, we sample element on the main diagonal of the above covariance matrices through Inverse Gamma priors. Finally, IBP parameter $\alpha$ is sampled through a conjugate Gamma($g,h$) prior \cite{knowles2011nonparametric}, as follows.

\begin{equation}
\label{eq:postSigma}
\begin{split}
& IG (\Psi_{y}(i)| a + {N \over 2}, b + tr(E_{y}(i)^T E_{y}(i))) \propto N(E_{y}(i)|0, \Psi_{y}(i)) IG(\Psi_{y}(i)|a,b) \\
& IG (\Psi_{z}(k)| a + {m_k \over 2}, b + tr(E_z(k)^T E_z(k))) \propto N(E_{z}(k)|0, \Psi_{z}(k)) IG(\Psi_{z}(k)|a,b) \\
& IG (\Psi_{q}(k)| c + {q \over 2}, d + tr(q_k q_k^T)) \propto N(q_k|0, \Psi_{q}(k)) IG(\Psi_{q}(k)|c,d) \\
& IG (\Psi_{p}(k)| c + {p \over 2}, d + tr(p_k^T p_k)) \propto N(p_k|0, \Psi_{p}(k)) IG(\Psi_{p}(k)|c,d) \\
& P (\alpha|S , e , f) \propto P(S|\alpha) P(\alpha|e,f) \propto G(\alpha| K + g, h + H_N)\\
\end{split}
\end{equation}

\section{Experiments}
Following the blueprint set forth in sections 1 and 4.1, we compare the models listed below. Please note that the first four NCFR variants use a simulated annealing regulariser, with $T_0 = 1000$ and cooling function $T = 0.9 T$. The zero Metropolis-Hastings alternative serves as a sparse model, bound to $K$'s initial value and only moving to lower ranks of the latent space. The code for NCFR and related models is inspired by \cite{KnowlesISFACode} and consists of configuration tools to run all the following models, soon to be available online.
\begin{enumerate}
\item Full-rank linear regression (FRR) 
\item Conditional Factor Regression (CFR)
	\begin{itemize}
		\item $K = 15$ (CFR15)
		\item $K = 20$ (CFR20)
		\item $K = 25$ (CFR25)
	\end{itemize}
\item Non-parametric Conditional Factor Regression (NCFR)
	\begin{itemize}
	\item Fixed $\alpha$ (Fix$\alpha$)
	\item Sampling $\alpha$ (Smp$\alpha$)
	\item Isotropic noise (Iso)
	\item Diagonal noise (Diag)
	\item Plain prior MH function (PMH)
	\item Simulated annealing MH function (SAMH) 
	\item Spike and slab MH function (SSMH)
	\item Zero MH function with $K_0 = 15$ (ZMH15)
	\item Zero MH function with $K_0 = 20$ (ZMH20)
	\item Zero MH function with $K_0 = 25$ (ZMH25)
	\end{itemize}
\end{enumerate}

\subsection{Synthetic data}
As an attempt to evaluate the models above, we have synthetically generated $X$ and $Y$ through an IBP-resembling process, since following the exact IBP steps would create bias towards the model. $Z$ is calculated as the product of a standard Gaussian random $X$ and zero-mean diagonal Gaussian $P$ with added Gaussian noise. It is then masked by a binomially-distributed $S$. The resulting sparse $Z$ is utilised to generate $Y$, along with $Q$ and diagonal Gaussian noise $\Psi_y$ (eq.~\ref{eq:model2}).

The synthetic data use a $70D$ input and a $50D$ response which are typical values in computer vision problems \cite{tenorth09dataset}. The latent factors are arbitrarily set to $20D$. We adopt two different training schemes. In the first, we randomly select 100 out of 1000 data as missing values and infer them iteratively with MCMC iterations. Alternatively, we train the models with 900 observations and use the remaining unseen 100 for test with the trained model. The parameters used are chosen as the sample with highest likelihood amongst the last 100. Both these methods yield similar accuracies, but the latter imposes far less computational cost for prediction, consisting of a single linear transformation.

Due to the unidentifiability of projection/load factor matrices in PPCA, factor analysis and similar models to NCFR \cite{Tipping99},\cite{west2003bayesian}, we have utilised prediction likelihood and least square error for performance measurement. Overfitting is monitored by auditing $K$ and comparing prediction accuracies in training and test. A final study on computational speed shows the practicality of our solution.

\subsection{Results}
\begin{figure}
\begin{center}
\includegraphics[scale =0.6,  clip]{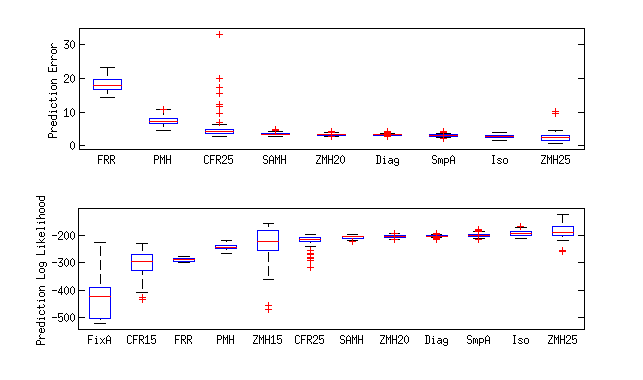}
\end{center}
   \caption{Boxplot of prediction performance over synthetic test data. (a) Normalised least square prediction errors are presented over dimensions of $Y$. Due to considerable range difference, the least accurate models are not shown here. (b) Prediction log-likelihoods of the last 100 MCMC samples. Log-likelihoods are sorted and aligned in accordance with prediction errors.}
\label{fig:SynpredBoxplot} 
\end{figure}

The main performance figure used in the experiment is prediction accuracy. Figure \ref{fig:SynpredBoxplot}(a) illustrates boxplots of normalised least square prediction errors over dimensions of $Y$. As a complementary measure of confidence, we plot prediction log likelihoods of test data over the last 100 iterations  (figure \ref{fig:SynpredBoxplot}(b)). Please note that both prediction error and likelihood values are sorted and aligned to illustrate their compatibility. As can be seen, the top 5 models are variants of NCFR isotropic and diagonal, with SA and zero ($K_0 = 25$) candidate functions. It is important to note that the SA alternative yields similar good results, but avoids the time consuming and arbitrary process of determining $K_0$. Spike and slab models do not prove useful in NCFR, resulting in heavy overfitting. The least accurate models including all CFRs and fixed $\alpha$ variant are not included in \ref{fig:SynpredBoxplot}(a) to preserve better visibility for the others.
\begin{figure}
\begin{center}
\includegraphics[scale = 0.6,  clip]{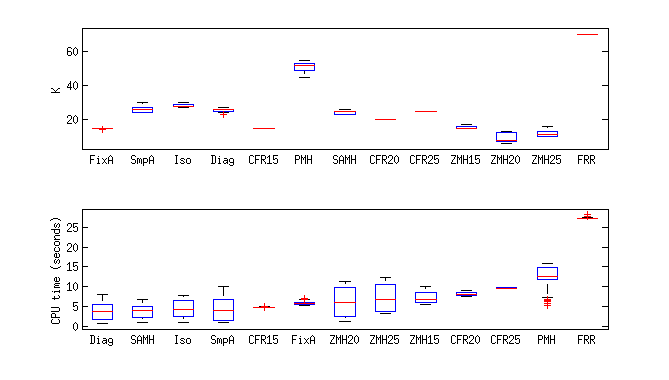}
\end{center}
   \caption{Boxplot of the number of reduced dimensions, $K$, and CPU time over the last 100 iterations. (a) Values of $K$: the closest to the original value, 20, are the NCFR variations with $K = 25$, showing a modest overfitting. (b) CPU time per iteration in seconds is sorted in ascending order. The two ends of the spectrum are the NCFR variants (best) and FRR (worst).}
\label{fig:KandCPUO}
\end{figure}
This prediction performance does not show the effectiveness of the IBP in deriving the expected dimension for $Z$. If the model is unable to fit the data with acceptable likelihood, the immediate consequence is resorting to more features in $Z$ to compensate for low likelihoods. Hence, we need to compare $K$ to the original simulation value, {\it 20}. Figure \ref{fig:KandCPUO}(a) shows that the best fit for $K$ is achieved by NCFR with only a modest over-estimate ($K = 25$). The greatest overfitting is provided by SSMH and PMH. It is worth mentioning that the differences between training and test prediction errors also support the above, with a mean delta of $0.113$ for default NCFR variations.  

Finally, we consider the computational costs NCFR to explore its viability. Figure \ref{fig:KandCPUO}(b) shows the CPU time per iteration for each model. In accordance with the above results, the NCFR variants (isotropic/diagonal noise and SA) are the fastest, even more efficient than CFRs, zero-MH variants and classic FFR. A way to explain this is through IBP's unique capacity in optimising $K$ for each observation and in each iteration. This can significantly improve speed particularly in high dimensions to surpass other parametric models (CFR and FFR). Additionally, it offers sparsity within each active factor which in turn causes more efficiency.

\section{Conclusion}

In this paper we have proposed a non-parametric solution for high-dimensional, multivariate linear regression. The proposed NCFR offers two improvements over classic linear regression. Through exploiting an intermediate latent space of lower dimensionality, NCFR reduces the degrees of freedom and mollifies overfitting. By integrating an IBP prior in the latent factor inference, it generates factors which are both sparse and unlimited in number; NCFR regularises this number by simulated annealing. Experimental results on a synthetic model with control of the number of actual dimensions and noise parameters prove that NCFR achieves remarkable prediction accuracy while maintaining a low computational cost, alongside a more accurate estimate of the latent dimensionality.

\bibliographystyle{unsrt}

\end{document}